\documentclass{llncs}
\usepackage[utf8]{inputenc}
\usepackage[T1]{fontenc}

\usepackage[english]{babel}
\usepackage{graphicx}
\usepackage{hyperref}
\usepackage{bm}
\usepackage{wrapfig}
\usepackage{amsmath}            
\usepackage{amssymb}            

\title{Multi-Sensor Data and Knowledge Fusion}
\subtitle{A Proposal for a Terminology Definition}
\author{Silvia Beddar-Wiesing, Maarten Bieshaar}
\institute{Intelligent Embedded Systems,
	University of Kassel, Germany\\
\email{\{s.beddarwiesing,mbieshaar\}@uni-kassel.de}}

\begin{document}
\maketitle
\begin{abstract}
	Fusion is a common tool for the analysis and utilization of available datasets and so an essential part of data mining and machine learning processes. However, a clear definition of the type of fusion is not always provided due to inconsistent literature. In the following, the process of fusion is defined depending on the fusion components and the abstraction level on which the fusion occurs. The focus in the first part of the paper at hand is on the clear definition of the terminology and the development of an appropriate ontology of the fusion components and the fusion level. In the second part, common fusion techniques are presented.
\end{abstract}
\section{Introduction}

The generic term \textit{fusion} describes the combination of different fusion components that consist of available datasets. A more specific definition of fusion is only possible regarding the context and the purpose of the fusion, i.e., in particular the components to fuse. In general, the assumption behind the application of fusion is, that  fusing datasets from different sources improves the performance of the subsequent data processing.

Consider the task of tracking a pedestrian at a crossroad with the help of a set of cameras. The aim is to generate a position prediction of a pedestrian at the next timestep.
For this purpose, predictions based on the images from different cameras can be combined to obtain a more robust prediction. The fusion on a previous stage of the data processing can improve the prediction accuracy. Fusing knowledge in form of models for predestrian behavior, fusing information about the location and velocity of the pedestrian or fusing images of the same pedestrian to obtain less noisy images can contribute to a more precise prediction.

Unfortunately, inconsistent vocabulary often complicates the clear definition and description of fusion algorithms and, as a result, makes the categorization of the datasets and the selection of a corresponding fusion technique difficult. The aim of the paper at hand is to clearify the terms of possible fusion components, the terms used for the different process levels and the definition of the total process. Several deceptive definitions are discussed and afterwards, an ordered definition for the fusion terms is proposed that combines the common definitions of the fusion components and the fusion level. In the last section, a selected set of fusion techniques for the different fusion level are listed.

\section{Subdivision of Fusion Techniques}

To define a categorization of fusion techniques, it is necessary to first define the terms used for the fusion components.
The definition of the fusion components corresponds to the level of abstraction that can be determined by means of the data-information-knowledge-wisdom (DIKW) hierarchy \cite{Ackoff1989}.

Furthermore, the constitution of the fusion components per level restrains the possible fusion algorithms to a specific family of techniques.
In the following, an extension of the DIKW hierarchy is illustrated and afterwards, the fusion levels are specified. 

\subsection{Fusion Components}\label{sectFusionComp}
There have been wide studies about the categorization of data. In this section, two popular concepts are examined considering the application of the data as fusion components.\\
One common categorization has been published by Ackoff \cite{Ackoff1989} and divides data into five categories that can be transferred into each other: data, information, knowledge, understanding, and wisdom.

Ackoff describes \textit{data} as representations of objects or events. Processing the data to improve the usability leads to \textit{information} that is used in descriptions and answers questions that begin with what, who, where, and how many.
The application of data and information generates \textit{knowledge} that can transform information into instructions and answers questions that begin with how.
If relations and patterns in the information are identified, the context has been captured and, as a result, \textit{understanding} has been reached. Understanding helps with questions that begin with why.
At least, \textit{wisdom} includes the ability of judgement and the competence of dealing with the value of the data, information and knowledge.
\begin{figure}[h!]
	\begin{center}
		\includegraphics*[scale = 0.5]{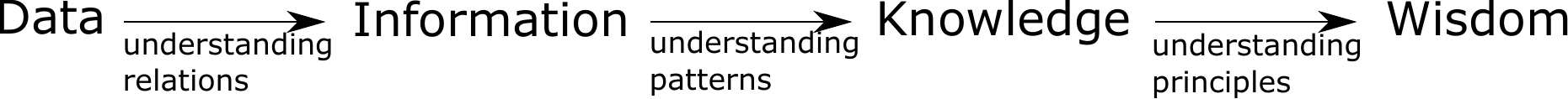}
		\caption{The adjusted DIKW hierarchy.\label{DIKW}}
	\end{center}
\end{figure}
\vspace*{-0.5cm}
According to the further elaboration of the definitions from Ackoff in \cite{bellinger2004data}, Bellinger et al. take the view that understanding is not a stage in the hierarchy but the condition for the transition from a lower to an upper level as schematized in Figure \ref{DIKW}. Thus understanding relations between data leads to information, understanding patterns in information generates knowledge, and understanding the underlying principles of knowledge results in wisdom.\\

With regard to the following fusion techniques that originate mostly from a machine learning context, the extended version of the DIKW-hierarchy provides a more applicable definition of the fusion components. In Section \ref{FusionTech}, the relation between the extended DIKW-hierarchy and the fusion levels will be clarified. But first, the fusion process and the levels of fusion are defined in the following.






\subsection{Definition of Fusion and Fusionlevel}
Especially in terms of fusion definitions, the literature varies.
In general, there exist two kinds of perspectives: On the one hand, there are universal definitions of fusion in sense of a whole process. On the other hand, fusion is performed on components from different abstraction levels. But due to the inconsistent classification of fusion techniques, the comparison of literature is often difficult. In particular, the definition of data and information fusion differs.

In the following sections, the terminology of different approaches is discussed and in the conclusion, an ontology of the terms for the fusion components and the fusion level is proposed in form of the fusion level rainbow as illustrated in Figure \ref{rainbow}.

\subsubsection*{Fusion.}
In \cite{Bostroem2007}, several differing previous definitions as well as a new definition of sensor, data and information fusion are listed.
Staying with the attempt to categorize the general definitions, it is noticeable, that they focus on three different aspects in particular:

Firstly, the \textit{fusion components} are a central aspect. In some literature, they are explicitly identified as observations or measurements respectively raw data \cite{Durrant-Whyte2012,McKendall1988,Waltz1990}. In most cases, they are referred to as data or information, or it is specified that they originate from multiple sources, but there is no detailed discussion about the representation of the data.\\

Secondly, some of the definitions also focus on the \textit{process} behind the fusion. A popular precedent for this is one of the first definitions of fusion published by the Joint Directors of Laboratories (JDL) in form of a data fusion lexicon \cite{white1986data}. The fusion process is described as the \glqq association, correlation, and combination of data and information from single and multiple sources\grqq \cite{white1986data}.
It is additionally mentioned, that the fusion has several levels that are explained separately as listed in the next section.
A different description of fusion in
\cite{Koch2016} interprets the process as the application of prior knowledge to a concrete realization.\\

Thirdly, most authors focus on the \textit{purpose} of the fusion or consider the combination of the three aspects presented. The most common goal that is stated includes an improvement of the obtained information due to fusion.
Further advantages are reflected in its applicability for many purposes:
The resulting understanding of the observed situation \cite{Varshney1997}, smoothed data and reduction in uncertainty \cite{Challa2005}, an optimal estimate of a hidden state \cite{Gao2009}, an improved perfomance of inference \cite{Hall1997}, improvement in prediction \cite{Steinberg2008} or decision tasks \cite{Fisch2014} or in general richer and more useful information \cite{Wald1999},\cite{Waltz1990}.

\subsubsection*{Fusionlevel.}
\begin{figure}
	\begin{center}
		\includegraphics*[scale=0.5]{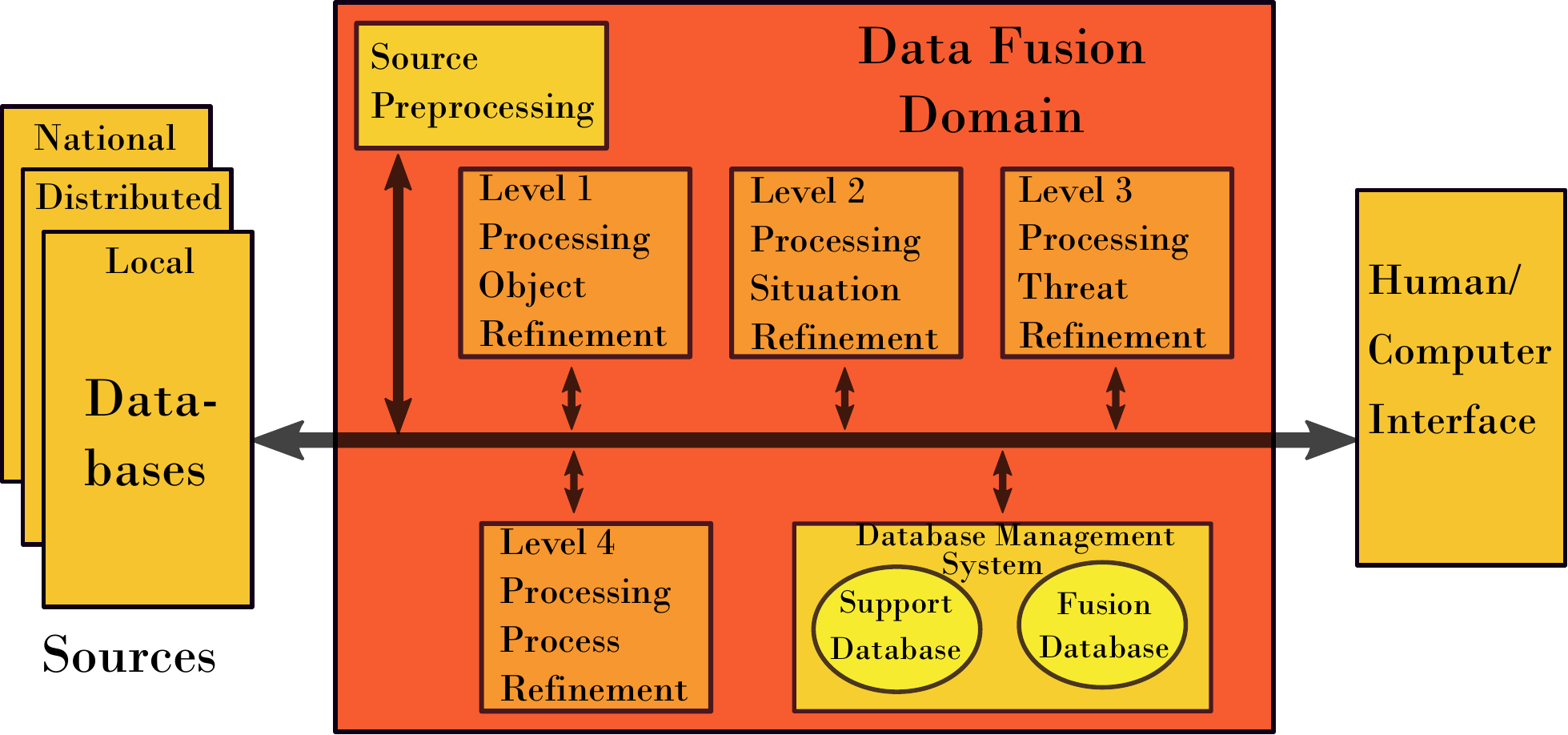}
		\caption[JDL model.]{The extended fusion level definition of the JDL includes six different levels that can be passed through during an entire fusion process (sketched here based on\cite{Hall2009}).\label{JDL}}
	\end{center}
\end{figure}

Many definitions of the level of fusion refer to the published data fusion lexicon of the JDL \cite{white1986data}. Here, the level specification originally has been analysed from the point of view of a military application, but it can be extended to several fusion applications. It consists of three interrelated levels, but it is more common to use the extended JDL definition. The latter is illustrated in Figure \ref{JDL} and includes the three levels from the JDL model (levels 1 - 3), extended by three more levels as described in \cite{Hall2009}. The six levels are defined as follows: \\

\begin{itemize}
	\item[\textit{Level 0:}] Fusion of raw data in form of signal refinement to obtain preliminary information about the characteristics of the observed object or situation.
	\item[\textit{Level 1:}]  Data is processed to specify the position or identity of an entity, or to classify characteristics of it. This is called the object refinement.
	\item[\textit{Level 2:}] Relationships between objects and events considering the environment leads to a situation refinement by, e.g., analyzing relation structures.
	\item[\textit{Level 3:}] Characterized as the threat refinement. In general, this can be interpreted as a risk estimation by drawing inferences or predictions for application-specific operations.
	\item[\textit{Level 4:}] The performance improvement of the entire fusion process by refining the elements of it during a suitable type of monitoring.
	\item[\textit{Level 5:}]  A process of cognitive refinement via optimizing the interaction of the process with the user, that is dissociated from the previous levels.
\end{itemize}

\vspace*{-0.8cm}
For a more differentiated categorization of fusion processes, the goal of the paper at hand is to combine the characterization of the fusion components from Section \ref{sectFusionComp} and the fusion levels (and eventually the constitution of the fusion components). But, from this point of view, the level definitions from the JDL are too broad and the interest of the type of fusion components emerges only in the first four levels. \\

In contrast to the JDL model, the model proposed by Dasarathy \cite{Dasarathy1997}, where fusion processes are divided into five levels, is more associated to the DIKW-hierarchy from Section \ref{sectFusionComp}. The reason is, that the choice of devision is directly related to the fusion components and the fusion emissions. Interpreting the knowledge as a feature of the observed event and wisdom as decisions, that can exist, e.g., as classification decisions, evaluation of a regression model or a prediction, the following model is directly based on the DIKW-hierarchy.

In addition, Varshney \cite{Varshney1997} added a sixth level, so that all possible ascending (relating to the components presented in \ref{sectFusionComp}) or equal pairs of inputs and outputs of a fusion are considered.
The all-encompassing extension of Dasarathy's model is illustrated in \cite{Hall2009} and will be briefly listed in the following. For this representation, the shortcuts DAI (input: data), DAO (output: data), FEI (input: feature), FEO (output: feature), DEI (input: decision) and DEO (output: decision) are used. All pairs of inputs and outputs during a fusion process can be interpreted as follows:

\begin{itemize}
	\item[1] DAI/DAO: signal detection (fusion of raw datasets generates data with less noise, a sharper signal can, e.g., be used to detect signal sections)
	\item[2] DAI/FEO: feature extraction (fusion of datasets generates data that can be used to extract relevant features)
	\item[3] DAI/DEO: Gestalt-based object characterization (fusion of raw datasets can lead to a better characterization of an object or a decision)
	\item[4] FEI/DAO: model-based detection and feature extraction (fusion of features leads to refined features from which data can be generated, e.g., the fusion of Gaussians and subsequent sampling)
	\item[5] FEI/FEO: feature refinement (fusion of different models that describe the same feature generates a more confident feature model)
	\item[6] FEI/DEO: feature-based object characterization (feature level fusion refines the description of an object or a decision)
	\item[7] DEI/DAO: model-based detection and estimation (decision fusion sharpens the decision model from which data can be generated e.g. mixture of generative experts and subsequent sampling with the Gibb's Sampler \cite{Bishop06})
	\item[8] DEI/FEO: model-based feature extraction (decision fusion can lead to a decision from which a feature can be derived)
	\item[9] DEI/DEO: object/decision refinement (e.g. mixture of experts leads to a better decision)
\end{itemize}

But in fact, the combination of data as input and decision as output does not often occur in common tasks. Additionally, cases 4, 7  and 8 are feature respectively decion level fusions from which can be sampled or conclusions drawn afterwards. Depending on the form of the feature representation, samples can be generated by common sampling methods as described in \cite{Bishop06}.

\newpage
Furthermore, there is a set of models that apply an additional backward connection from the top level to the data acquisition process as in the OODA (Observe, Orient, Decide, Act) loop \cite{Bedworth2000}. As a result, the generation and collection process of input data can be adapted with regard to the conclusive decision evaluation.

\begin{figure}[h!]
	\begin{center}
		\includegraphics*[scale=0.5]{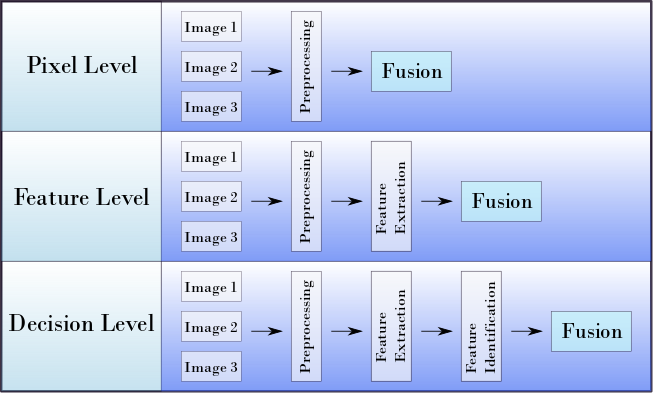}
		\caption{For image data, the data level fusion is known as pixel level fusion and used for image processing (figure inspired by the depiction from \cite{Chang2018}).}\label{pixel}
	\end{center}
\end{figure}

One last model has to be mentioned here that is also common, but especially used in image fusion applications and is similar to the described definitions before. Here the fusion is divided into the \textit{pixel, feature} and \textit{decision level} \cite{Chang2018}.
Analogous to the data, feature and decision fusion, this model is applied to image processing, which is a subset of the sensor fusion. An illustration is given in Figure \ref{pixel}.

In order to combine the delineated fusion level terms, the following ordered ontology describes the relations between the terms:\\

\begin{figure}[h!]
	\begin{center}
		\includegraphics[width=4.5cm,height=10cm]{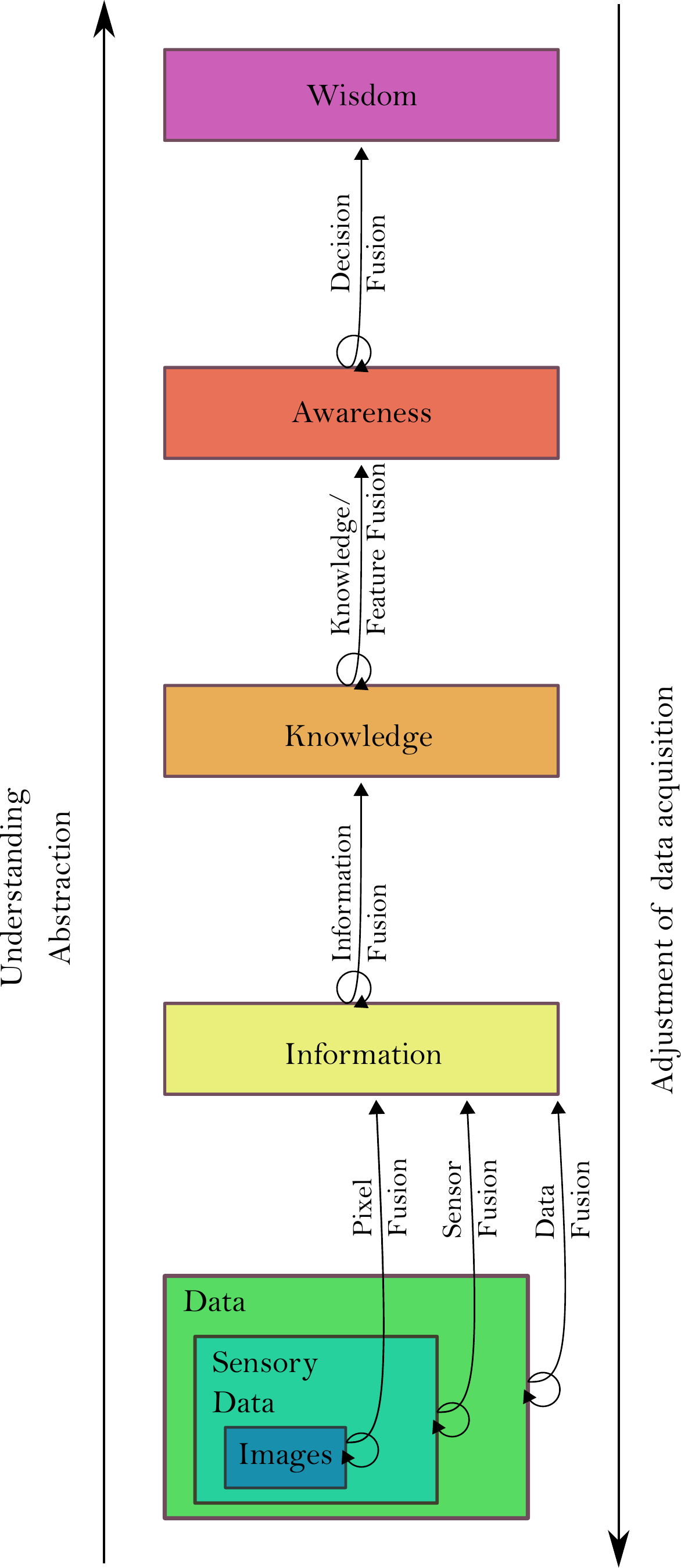}
		\caption{The Fusion Level Rainbow. Based on the extended Dasarathy model, the fusion within a component leads to a refined element of the corresponding component or to a more abstract emission.\label{rainbow}}
	\end{center}
\end{figure}

\begin{itemize}
	\item[$\circ$] (Pixel, Sensor) Data Fusion
	\item[$\circ$] Information Fusion
	\item[$\circ$] Knowledge/Feature Fusion
	\item[$\circ$] Decision Fusion
\end{itemize}
and is illustrated in Figure \ref{rainbow}.

\newpage
At the bottom of the entire fusion process, raw data given by sensors or other sources are fused while understanding characteristics and relations of the input. The refined data with a low information loss compared to the original data provides an updated representation for further applications.

The obtained information is a fundamental basis for the feature extraction in the next level that proposes an underlying model for the data. This can provide knowledge that can be used to understand patterns in the data and thus create awareness of underlying principles in the source data.

On top of the fusion process, the aim is to gain wisdom in form of  performance improvement in decision making and thus the choice of action. Depending on the impact of the action, the entire fusion process can be adapted at the different stages.

\newpage
\section{Fusion Techniques of the Particular Levels}\label{FusionTech}
In this section, selected techniques from different fusion levels will be presented. At the beginning of \ref{SensorFusion}, a fine structure of sensor fusion techniques is outlined. Afterwards, statistical fusion methods are described in more detail.
To get a short overview of algorithms used in data and information fusion, the focus in Section \ref{DataInfoFusion} is on the constitutions of data that require different fusion processes.
At the end in Section \ref{KnowFusion} and \ref{DecFusion}, different forms of knowledge and decision fusion techniques are characterized.
\subsection{Sensor Fusion}\label{SensorFusion}
In the fusion level rainbow (Fig. \ref{rainbow}), the sensor fusion is a subset of the lowest level of fusion, the data fusion. Sensor data is a special case of data that can be represented as a data point in a high dimensional space and is produced by (multiple) sensors.
Sensor fusion techniques once again can be categorized according to the information flow between the available sensor network and different sensor configurations.\\

Firstly, the fusion can be implemented \textit{centralized} or \textit{decentralized} \cite{Gustafsson2012}. In the centralized architecture, the measurements of all sensors are available during the fusion process, so a batch method is used. In contrast to this, in the decentralized fusion, the measurements of each sensor is fused within a seperate fusion model. Then during the global fusion process, only the model information of each sensor is available and processed sequential. The decentralized fusion is preferred since the fusion process is considered as being more robust and reliable \cite{Gao2009}.

Secondly, sensor fusion can be furthermore divided into three cases depending on the sensor configuration as listed in \cite{Durrant-Whyte1988}:
\subsubsection*{Competitive Sensor Fusion.} (homogeneous)
Either data from sensors of the same modality are fused or the sensors can be transformed to the same baseline previously and are fused afterwards. Data fusion of competitive sensors can be used to reduce noise respectively uncertainty.

In connection to the initial example, competitive sensors in form of cameras used for the pedestrian tracking produce images of the same person at the same time. Fusing the images that may contain a degree of uncertainty, the resulting images are less noisy and more applicable for the tracking task.
\subsubsection*{Complementary Sensor Fusion.} (heterogeneous)
Sensors observe the same event and fusion of them generates a complemented image of the observation. This means, the sensors can measure different and disjunct parts of the same event and the combination leads to a complete characterization of it.

A complementary set of cameras, e.g., can provide an extended picture of a crossroad in contrast of the image of only one camera which simplifies the subsequent tracking of a pedestrian.
\subsubsection*{Cooperative Sensor Fusion.}
A sensor is configurated depending on the information from other sensors to generate more useful information. 
This form of sensor network configuration includes some sort of temporal delay and dependency on a decision of an expert. The tracking task can require the possibility to adapt the camera angles after observing a certain behaviour of the pedestrian.\\


Sensor fusion techniques also differ in the assumptions about the system under consideration. For processing sensor data with uncertainties, statistical sensor fusion techniques for static and dynamic systems are presented by Fredrik Gustafsson in \cite{Gustafsson2012}:
\subsubsection*{Statistical Sensor Fusion}
The main idea behind statistical models is, that the sensors are noisy and the \glqq true\grqq \ characterization of the event is given by a state vector $\bm{x}$. Furthermore, it is expected, that by fusing different sensors, the state vector gets more precise.

The assumption here is that $N$ given observations $\bm{y}_n\in\mathbb{R}^{n_y},\ n\in [ N]$, stacked in $\bm{y}\in\mathbb{R}^{N\times n_y}$, can be described by a model
\begin{align*}
\bm{y} = \bm{h}(\bm{x}) + \bm{e}.
\end{align*}
Gustafsson discusses linear models $\bm{h}(\bm{x}) = \bm{Hx}$ with a stacked factormatrix $\bm{H}\in\mathbb{R}^{N\cdot n_y\times n_x}$, as well as nonlinear models, that relate the observations to the hidden state vector $\bm{x}\in\mathbb{R}^{n_x}$. Most of the time, the stacked error $\bm{e}\in\mathbb{R}^{N\cdot n_y}$ is assumed to be Gaussian, but the non-Gaussian case is mentioned too.

In the \textit{static case}, the hidden state $\bm{x}\in\mathbb{R}^{n_x}$ is time-invariant, so that the model describes, e.g., the observation of the same event by means of several sensors or of the same sensor at different timestamps.

In the \textit{dynamic case}, the state $\bm{x}_k\in\mathbb{R}^{n_x}$ is additionally varying with time according to a sequential update model
\begin{align*}
\bm{x}_{k+1} = \bm{f}(\bm{x}_k) + \bm{v}_k,
\end{align*}
with a linear or non-linear mapping $\bm{f}:\mathbb{R}^{n_x}\rightarrow\mathbb{R}^{n_x}$ and a Gaussian or non-Gaussian noise $\bm{v}_k\in\mathbb{R}^{n_x}$.

For providing a new state estimation, Gustafsson gives an insight into a variety of least squares approaches that can be applied, while the measurements are independent.

In case of correlated measurements in the static case, Gustafsson presents the \textit{safe fusion algorithm}, which will be discussed in \ref{corrData}.

For dynamic systems, Gustafsson lists popular filtering algorithms, such as several variations of the Kalman Filter \cite{Kalman1960}, that make inference on the state from the observations using dynamic linear or nonlinear models. Numerical methods approximating the nonlinear filter models are also discussed. In the grid-based methods, e.g., parts of the calculation during the filtering process are approximated by discretising the state space \cite{Jazwinski2007} or replacing integrals by finite sums \cite{Kramer1988}.

\subsection{Data and Information Fusion}\label{DataInfoFusion}
In this section, the case of data respectively information fusion in the second lowest levels of the fusion level rainbow (Fig. \ref{rainbow}) in general are under consideration.
When selecting a data fusion technique, the first aspect to focus on is the constitution of the data. In \cite{KHALEGHI201328}, a detailed overview of fusion techniques for the following types of data quality is given:

\subsubsection*{Imperfect Data.} The case of imperfect data occurs in nearly all applications of data fusion. To deal with a certain degree of imperfect data, the following algorithms, that are described together with former extensions in \cite{KHALEGHI201328}, can be used:
\begin{itemize}
	\item \underline{\textit{Probabilistic Fusion:}} The main idea is to represent the imperfection of data by means of uncertainty in form of probability distributions. Fusing them according to the Bayesian fusion formula from \cite{Bishop06}:
	\begin{align}\label{BayesFusion}
	p(\bm{X}|\bm{Z}) = \frac{p(\bm{Z}|\bm{X})\cdot p(\bm{X})}{p(\bm{Z})}
	\end{align}
	determines the posterior probability distribution of the (real, underlying) state $\bm{X}$ depending on the observations $\bm{Z}=\lbrace\bm{z}_1,\ldots,\bm{z}_t\rbrace$, the likelihood $p(\bm{Z}|\bm{X})$ and the (chosen) prior distribution $p(\bm{X})$. Analogously, the Bayesian fusion can be formulated for the dynamic case by using observations up to time $t$ \cite{Gustafsson2012}.

	Based on this Bayes theorem, several fusion techniques have been formulated, that are used in algorithms for numerous applications, such as the Kalman filter. For more information see \cite{KHALEGHI201328}. 
	\item \underline{\textit{Evidential Belief Reasoning:}} In the Dempster-Shafer evidential theory (DSET), possible measurement hypotheses obtain correspoding beliefs and plausibilities. Usually, Dempster's rule of combination is used to fuse two belief masses of the same set by summing up the product of the belief masses of every pair of supersets regarding conflicting subsets.
	Consequential, the DSET provides a fusion technique that can be seen as a generalization of the Bayesian fusion where probability mass functions are used as belief functions \cite{Shafer1976}.\\
	The DSET has been proposed to represent incomplete data, analogously to the probability theory by modeling the membership uncertainty of an element in a well-defined class.
	\item \underline{\textit{Fusion and Fuzzy Reasoning:}} In contrast, the fuzzy set theory is mainly designed to represent and to operate on vague data and to model the fuzzy  membership of an element in an ill-defined class \cite{KHALEGHI201328}. For this purpose, a gradual membership function is introduced 
	that defines a fuzzy set 
	by assigning a membership degree between 0 and 1 to each element of the (discrete) universe. 
	The higher the degree, the more the element belongs to the fuzzy set. \\
	Fusing of membership degrees can be done in form of conjunctive and disjunctive fusion rules to obtain a fuzzy fusion output (i.e., the fusion function is bounded from above by the minimum in the former and from below by the maximum in the latter case), as further described in \cite{zadeh1965fuzzy}.
	\item \underline{\textit{Possibilistic Fusion:}} Based on the fuzzy set theory, possibility theory is conceived to again represent incomplete data by modeling an uncertain membership of the elements in the universe in well-defined classes with a possibility distribution \cite{KHALEGHI201328}. Another difference to fuzzy theory is to require normalized membership functions that can be used to define a possibility and a necessity degree. The former determines the plausibility and the latter the certainty of a subset of the universe. So the possibility theory differs from probability theory by the use of the two dual set-functions.
	The fusion rules are identical to the ones for the fuzzy fusion.
	\item \underline{\textit{Rough Set Based Fusion:}} The idea of the rough set theory is to approximate a data set by an upper and lower bound of sets to obtain a rough representation of the original set. The lower bound set includes subsets of the data that definitely belong to the original set. The difference of the upper and lower bound sets includes subsets that cannot be classified as belonging or not belonging to the original set. In the approximation, the granularity of the data can be considered by choosing an appropriate size of the united subsets. \\
	Datasets can then be fused by combining the rough sets of their approximations with the aid of classical conjunctive or disjunctive fusion techniques of set theory such as the union or intersection of sets \cite{KHALEGHI201328}.
	\item \underline{\textit{Random Set Theory:}\label{randomSetTheory}} The theory of random sets uses a form of generalized random variables, the random sets, which take sets as random values. In \cite{Goodman2013}, random sets and their characteristics are described in detail. In addition, Goodman et al. discuss applications of random sets in single- and multi-target data fusion which can be used for the data association problem. \\
	The handling with coarse data in general by using the random set theory is described in \cite{Nguyen2006} extensively.

\end{itemize}

\subsubsection*{Correlated Data.}\label{corrData}
As mentioned in Section \ref{SensorFusion}, correlated data can be fused by the \textit{safe fusion algorithm} \cite{Gustafsson2012} or evolved algorithms as listed in \cite{KHALEGHI201328}. Here it is assumed, that the correlation is unknown and the data respectively information is represented as a point in a high dimensional space.

\begin{figure}[h!]
	\begin{center}
		\includegraphics*[width=\linewidth]{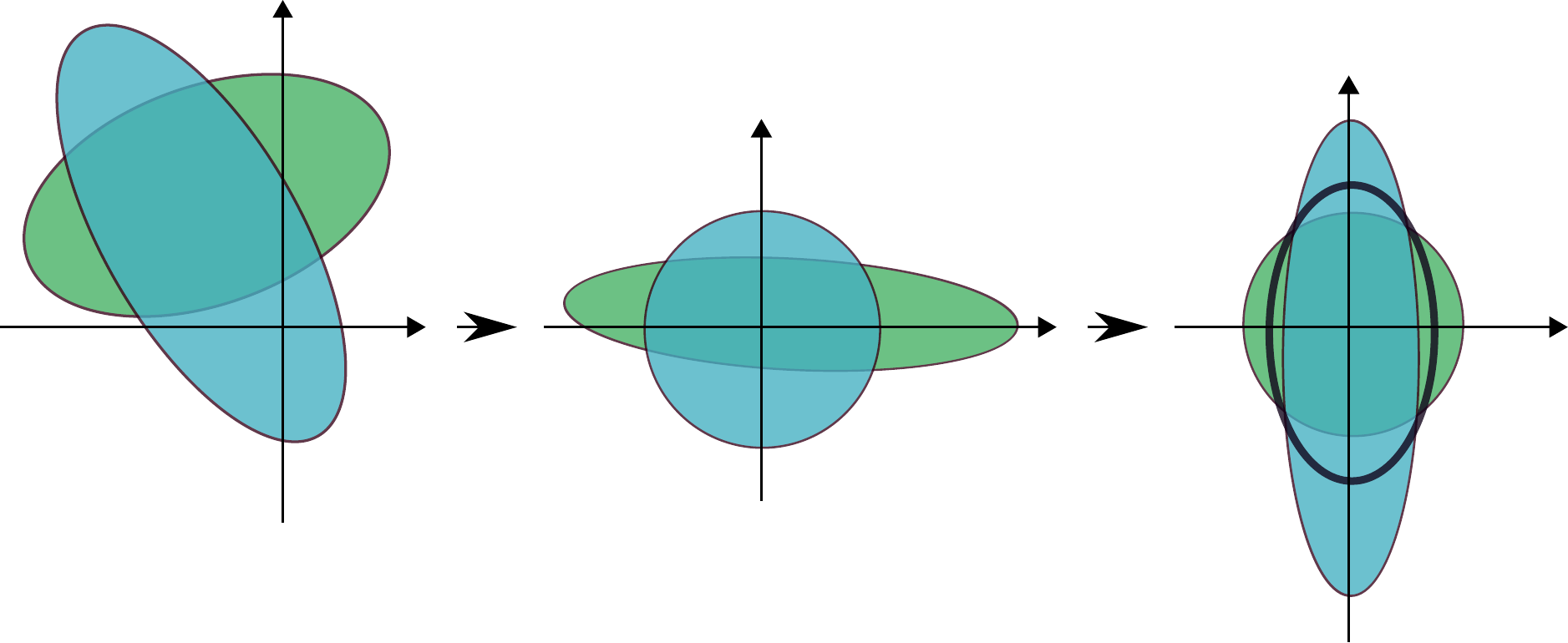}
	\end{center}
	\caption{Safe fusion algorithm visualized (similar as illustrated in \cite{Gustafsson2012}) in the fusion of two covariances. 
		In the first step, both covariances are transformed corresponding to the SVD of one covariance (blue). The SVD of the obtained manipulated second covariance (green) yields in the second transformation of both covariances. The final intersection of the covariances is then covered by the target covariance (thick ellipse). \label{safeFusion}}
\end{figure}

The safe fusion algorithm presumes a decentralized network and is based on the covariance intersection (CI) algorithm proposed by Julier and Uhlman \cite{Julier1997}. This means in addition, it is assumed that the models of the depending measurements include Gaussian noise and thus possess individual means and covariances.

The algorithm determines a covariance that defines the minimal ellipsoid that encloses the intersection of (sequentially entering) covariances by making use of a transformation defined by the sequentially emerging singular value decomposition (SVD, \cite{Golub1970}) as illustrated in Figure \ref{safeFusion}.

In case of a known correlation, it can be eliminated before fusing the data \cite{KHALEGHI201328}, e.g., by the principal component analysis \cite{Bishop06}.
\newpage
\subsubsection*{Inconsistent Data.} There are several forms of inconsistencies, which must be dealt with in different ways. These are listed in the following:
\begin{itemize}
	\item \underline{\textit{spurious data:}} Sensor data can be distorted by permanent failures or slowly evolving errors of a sensor. The most common ideas of dealing with spurious data is the identification or prediction of systematic errors. After that, suspicious data can be excluded from the fusion process. A statistical approach for detecting spurious data can be found in \cite{Kumar2007}. Kumar et al. formulated an extension of the basic Bayesian fusion formula \eqref{BayesFusion} by augmenting a random variable which describes whether the data is spurious or not depending on the data and the true state.
	\item \underline{\textit{out-of-sequence-measurements (OOSM):}}
	There are two different aspects that have to be considered in dynamic systems: the validity and the out-of-sequence arrival of data. When a history of measurements is necessary for the fusion process, the data should be updated at appropriate intervals to guarantee a valid fusion result. In addition, an implementation of an entire sensor network can entail delayed data deliveries for the fusion process. A possible approach for these problems has been proposed by Kaugerand et al. in \cite{kaugerand2018time}, that defines a fusion interval in which incoming data is permitted to be fused. Also in \cite{KHALEGHI201328}, some strategies for dealing with out-of-sequence data are presented.
	\item \underline{\textit{conflicting data:}} The problem of conflicting data results in misleading conclusions as discussed in \cite{Zadeh1984}. Extensions of the DSET have been developed to especially address the problem of inconsistent data and are mentioned in \cite{KHALEGHI201328}. \\

	In a statistical environment, in complement to the CI algorithm, the covariance union (CU) algorithm can be applied to deviating measurements for a consistent fusion with the assumption of a Gaussian uncertainty \cite{Uhlmann2003}. The union process consists of the computation of a mean and a covariance, so that the previous covariances added to the deviation of the previous mean from the new mean constitute a lower limit of e.g. the determinant of the new covariance.
\end{itemize}

\subsubsection*{Disparate Data.}  An aspect, that has not been considered yet is the fusion of data from sensors and sensor intelligence (hard information), data from human intelligence, open source intelligence and communications intelligence (soft information) \cite{Pravia2008}. Hard information can be represented in a mathematical framework and can therefore be used for the fusion techniques as presented above. \\
Whereas soft information is produced by human sources and is therefore available in \glqq context-dependent
languages  over  bandwidth-limited  channels\grqq \;\cite{Pravia2008}. The research on modeling uncertainty of such soft information is quite young, but there are some models for linguistic data described in \cite{Auger2008}. \\

\subsection{Knowledge Fusion}\label{KnowFusion}
When the abstraction of the fusion components increases along the fusion level rainbow (Fig. \ref{rainbow}), the components can be available in form of models that include knowledge from the observed event.

Knowledge fusion itself has two different levels: it can be performed at model or parameter level.

\subsubsection*{Model Fusion.}

Knowledge can be represented in form of different models. A simple example of such a model is a Gaussian distribution that includes information about the distribution of the data \cite{Bishop06}. In addition, artificial data can be generated with it. This kind of model is called a generative model.

To fuse knowledge from several models in form of \textit{mixture models}, they have to be in the same modality \cite{Bishop06}, i.e., they consist of the same base model but are trained differently or model different aspects of the underlying data. A high diversity in knowledge is prefered for a fusion as discussed for the case of ensembles in \cite{Ren2016}. In case of fusing models, the training can differ in the training set or the prior parameter setting.

The fused knowledge is then composed of a linear combination of the distinct models.
The mixture of Gaussian distributions, e.g., results in a Gaussian mixture model (GMM). It descibes the distribution of a data set, that is assumed to be more complex than a unimodal Gaussian, by a convex combination of a selected number of Gaussians.
The concluding mixture model contains more precise knowledge about the overall distribution and the mixture coefficients imply additive knowledge about the responsibility of a mixture component for generating a given data point.

Further examples for model fusion techniques are Convolutional Neural Networks \cite{Bishop06} and Multiple Kernel Learning based Ensemble Methods \cite{Goenen2011}. The latter approach permits different kernels representing knowledge that are fused by a (non-) linear combination function.

\subsubsection*{Parameter Fusion.}

Moreover, knowledge given by classification models can be available as classification rules or decisions (as the outputs of the classifiers). The combination of classifiers at the component level is equivalent to the mixture models and the combination at the output level will be presented in the next section.

The parameter fusion is a more complex fusion form and is applied on the parameter level of the classification models. In \cite{Fisch2014}, a knowledge fusion technique for generative classifiers based on mixture models (CMMs) is presented.

For the fusion algorithm, two or more probabilistic generative classifier that describe the same process have to be given. Each one is divided by the number of classes into parts that are mixtures of probabilty densities conditioned by the class and the mixture component. Additionally, the densities have to be defined on the same input space, i.e., especially on the same number of continuous and categorical dimensions and are trained on different training sets.

Before applying the algorithm, conjugate hyperdistributions for all hyperparameters per component are introduced and trained via the variational inference (VI) algorithm as described in \cite{Fisch2014}.

Having defined the hyperdistributions, the first step of the algorithm determines similar hyperdistributions of each component via an appropriate similarity measure. Assuming that a posterior distribution for a class is calculated by the Bayesian formula \eqref{BayesFusion} and the classifiers use the same prior knowledge, the fusion rule of two similar hyperdistributions is determined by the multiplication of the two posteriors divided by the prior. The derivation of this rule can be found in \cite{Fisch2014}.

Using conjugate priors, the fused posterior distribution has the same functional form as the given classifiers, so the estimation of the fused parameters are deducible from a sequence of mathematical transformations of the fusion rule.

For setting the distributions in the continuous dimensions to multivariate Gaussians and the distributions of the categorical ones to multinomial distributions, the hyperdistributions are Dirichlet respectively normal-Wishart distributions. The fusion formulae for the corresponding parameters are listed in \cite{Fisch2014}.

\newpage
\subsection{Decision Fusion}\label{DecFusion}
The decision fusion of multiple classifiers may consist of the direct combination of decisions or the selection of one suitable classifier for a specific input area. 
\begin{itemize}
	\item \underline{\textit{Committee/Ensemble}}: The decision is given as a combination 
	of decisions from distinct models \cite{Bishop06}. Boosting is an ensemble technique that trains the models iteratively and a decision is calculated by, e.g., weighting the models depending on their performance as in the adaptive boosting (AdaBoost) algorithm \cite{Freund1997}.
	\item \underline{\textit{Decomposition-Based Ensemble Methods:}} In case of time series data, several ensemble methods based on the lossless  decomposition of the input signal can be used for forecasting \cite{Ren2016}, which can be interpreted as decision making. After decomposing the time series into a set of signals that fully represent the original signal, the predictions from the components can be combined to one predition for the origin.
	\item \underline{\textit{Mixture of Experts:}} Depending on the input domain, a decision is made from one (hard or soft) selected model from the mixture of experts. The selection can be implemented in form of decision trees. Whereas in the Bayesian Model Averaging, one model is determined for the entire input space by introducing a prior probability for each model \cite{Bishop06}.
\end{itemize}

%

\section{Conclusion}
The ideal definition should have no space for interpretation or double allocation of the terminology. The goal of this paper was to provide a clear definition for fusion and its components for the applicability in varous contexts.
After having discussed several definitions, the fusion levels have been defined considering the previous specified fusion components. The resulting fusion level rainbow (Fig. \ref{rainbow}) includes the entire range of fusion components and levels and provides a clear definition of fusion for further applications. Finally, an overview of common fusion techniques was given for each of the previously defined levels.

In the presented definitions, decisions are often associated with predictions or classifications. The definitions are analogously valid for regression problems.

The concurrent fusion of components from different levels has not been considered in the paper at hand and is left for future work. In addition, the hybridization of the fusion techniques are in common use and have not been addressed here. 

\section*{\large Acknowledgment}

This work results from the project DeCoInt$^2$, supported by the German Research Foundation (DFG) within the priority program SPP 1835: "Kooperativ interagierende Automobile", grant number SI 674/11-1.

\newpage
\bibliographystyle{abbrvdin}
\bibliography{myBibliography}


\end{document}